\documentclass[letterpaper, 10 pt, conference]{ieeeconf}  % Comment this line out if you need a4paper
\IEEEoverridecommandlockouts

\usepackage[utf8]{inputenc} % usually not needed (loaded by default)
\usepackage[T1]{fontenc}
\usepackage{amsmath}
\usepackage{amsfonts}
\usepackage{float}
\usepackage{graphicx}
\usepackage{booktabs}
\usepackage[font=small]{caption}
\usepackage{subcaption}
\usepackage[ruled]{algorithm2e}
\usepackage{booktabs} 
\usepackage{enumerate}
\usepackage{times}
\usepackage{soul}
\usepackage{url}
\usepackage{hyperref}
\usepackage{bm}
\usepackage{adjustbox}
\usepackage{xcolor}
\usepackage{placeins}
\usepackage[utf8]{inputenc}
\usepackage{booktabs}
\usepackage{array, multirow}
\usepackage[capitalise]{cleveref}
\usepackage{tikz}
\usepackage{wrapfig}
\usepackage{adjustbox}
\usepackage{multirow}
\usepackage{amssymb}% http://ctan.org/pkg/amssymb
\usepackage{pifont}
\usepackage{algorithmic}
\usepackage{cleveref}
\usepackage{mathabx}
\begin{document}

\newcommand\mycommfont[1]{\footnotesize\rmfamily\textcolor{blue}{#1}}
\usetikzlibrary{arrows.meta}
\usetikzlibrary{positioning}
\tikzstyle{decision} = [diamond, draw, fill=blue!20, 
    text width=6em, text badly centered, node distance=3cm, inner sep=0pt]
\tikzstyle{block} = [rectangle, draw, fill=gray!10, 
    text width=10em, very thick, text centered, rounded corners, minimum height=2.2em]
\tikzstyle{line} = [draw, -{latex[scale=15.0]}]
\tikzstyle{cloud} = [draw, ellipse,fill=red!20, node distance=3cm,
    minimum height=2em]
\setlength{\fboxrule}{1pt}
\setlength{\fboxsep}{0pt}  

\newcounter{task}
\newcommand{\task}[2]{%
  \refstepcounter{task}%
  \label{#1}%
  \textit{Task~\thetask\ -- #2}%
}

\newcommand{\red}[1]{\textcolor{black}{#1}}
\newcommand{\green}[1]{\textcolor{green}{#1}}
\newcommand{\blue}[1]{\textcolor{blue}{#1}}

\captionsetup{skip=1pt}
\setlength{\textfloatsep}{1pt}
\setlength{\belowdisplayskip}{1pt} \setlength{\belowdisplayshortskip}{1pt}
\setlength{\abovedisplayskip}{1pt} \setlength{\abovedisplayshortskip}{1pt}
\setlength{\floatsep}{1pt} \setlength{\textfloatsep}{1pt}
\setlength{\intextsep}{1pt}
\setlength{\abovecaptionskip}{1pt}
\setlength{\belowcaptionskip}{1pt}
\captionsetup{belowskip=1pt}

\newcommand{\cmark}{\ding{51}}%
\newcommand{\xmark}{\ding{55}}%

\newtheorem{innercustomthm}{Theorem}
\newenvironment{customthm}[1]
  {\renewcommand\theinnercustomthm{#1}\innercustomthm}
  {\endinnercustomthm}

\newtheorem{innercustomprop}{Proposition}
\newenvironment{customprop}[1]
  {\renewcommand\theinnercustomprop{#1}\innercustomprop}
  {\endinnercustomprop}

\newtheorem{definition}{Definition}[section]
\newtheorem{prop}{Proposition}[section]
\newtheorem{theorem}{Theorem}[section]
  
\captionsetup[figure]{size=small}

\title{\LARGE \bf Affordance Transfer Across Object Instances via \\ Semantically Anchored Functional Map
}

\author{Xiaoxiang Dong$^{1}$ \and Weiming Zhi$^{2,3,4}$
\thanks{$^{1}$ Robotics Institute, Carnegie Mellon University, Pittsburgh, PA, USA.}
\thanks{$^{2}$ School of Computer Science, The University of Sydney, Australia.}
\thanks{$^{3}$ Australian Centre for Robotics, The University of Sydney, Australia.}
\thanks{$^{4}$ College of Connected Computing, Vanderbilt University, TN, USA.}
% %
% %Use only for final RAL version
% %\thanks{
% %Digital Object Identifier
% %(DOI): see top of this page.} 
}
% \author{Anonymous Authors}
\maketitle

\begin{abstract}
Traditional learning from demonstration (LfD) generally demands a cumbersome collection of physical demonstrations, which can be time-consuming and challenging to scale. Recent advances show that robots can instead learn from human videos by extracting interaction cues without direct robot involvement. However, a fundamental challenge remains: how to generalize demonstrated interactions across different object instances that share similar functionality but vary significantly in geometry. In this work, we propose \emph{Semantic Anchored Functional Maps} (SemFM), a framework for transferring affordances across objects from a single visual demonstration. Starting from a coarse mesh reconstructed from an image, our method identifies semantically corresponding functional regions between objects, selects mutually exclusive semantic anchors, and propagates these constraints over the surface using a functional map to obtain a dense, semantically consistent correspondence. This enables demonstrated interaction regions to be transferred across geometrically diverse objects in a lightweight and interpretable manner. Experiments on synthetic object categories and real-world robotic manipulation tasks show that our approach enables accurate affordance transfer with modest computational cost, making it well-suited for practical robotic perception-to-action pipelines.
\end{abstract}

% \begin{IEEEkeywords}
% Motion Generation, Dynamical systems
% \end{IEEEkeywords}

\section{Introduction}

Learning from demonstration (LfD) and imitation learning have long served as fundamental paradigms for teaching robots manipulation skills without manually specifying controllers. Traditionally, such approaches require robots to physically execute tasks during data collection, which limits scalability and makes acquiring diverse demonstrations time-consuming. Recent advances \cite{dong2025jointflowtrajectoryoptimization} show that robots can instead learn from human videos, from the same environment as the robot. This is achieved by extracting interaction cues with the identified objects without direct robot involvement, greatly expanding the availability of demonstrations. 

A key challenge that emerges in this setting is \emph{object generalization}: a robot should be able to observe an interaction on one object and transfer it to other objects that share similar functionality but differ in geometry. Humans naturally achieve this by first identifying corresponding functional regions between objects, such as handles, rims, or graspable parts, and then adapting their actions across the remaining structure. This observation suggests that effective generalization requires an explicit mechanism to align semantically meaningful object regions while maintaining smooth correspondence over the object surface. 

\begin{figure}[t]
    \centering
    \includegraphics[width=\columnwidth]{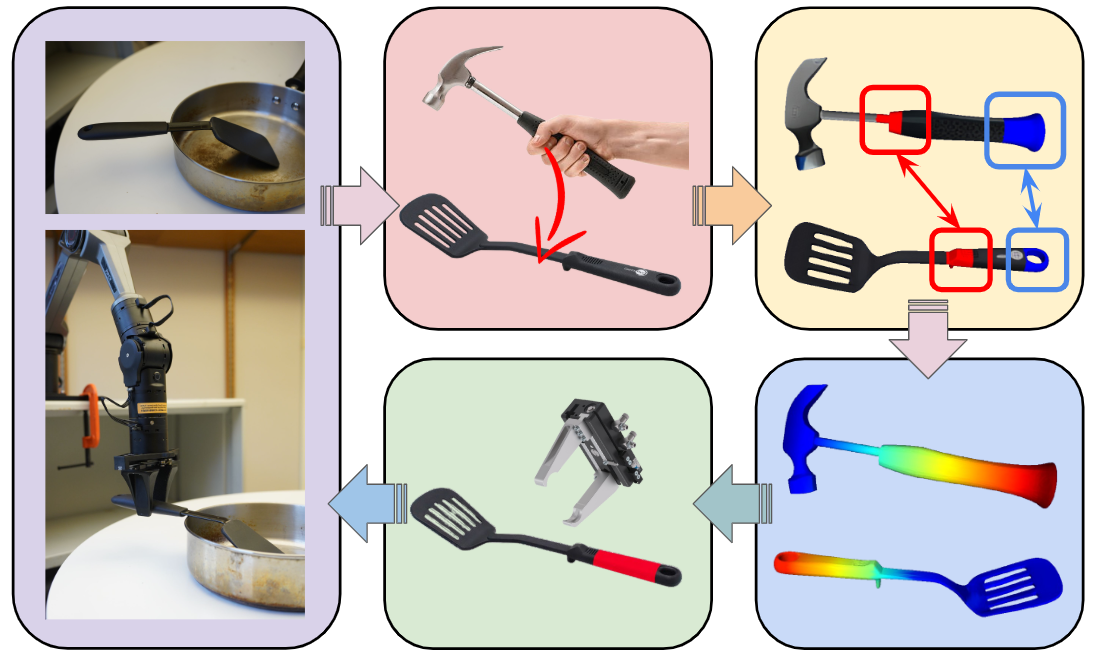}
    \caption{Overview of the proposed Semantic Anchored Functional Map pipeline. Starting from a single RGB observation of a demonstrated interaction, we reconstruct a coarse object mesh and extract the demonstrated affordance region. Semantic features are used to identify corresponding anchor regions across objects, which constrain a functional map to produce a smooth dense correspondence. The transferred affordance region on the target object is then used to generate a feasible grasp, enabling execution on a real robot.}
    \label{fig:figure1}
\end{figure}

In this work, we introduce \emph{Semantic Anchored Functional Maps} (SemFM), a framework that transfers affordances across objects by anchoring correspondence at semantically meaningful regions and propagating these constraints smoothly over the object surface using functional maps. By operating in a reduced functional space and leveraging intrinsic surface structure, our method avoids the computational overhead of VLM-based pipelines while retaining strong semantic alignment.

An overview of the pipeline is shown in ~\cref{fig:figure1}. This pipeline enables accurate affordance transfer across geometrically diverse objects while remaining computationally efficient, making it well-suited for robotic applications that require fast perception-to-action loops. Through extensive evaluation on synthetic categories and real-world robotic experiments, we show that our method achieves affordance transfer accuracy comparable to multi-view VLM-based approaches while being significantly more efficient.

In summary, this paper makes the following contributions:
\begin{itemize}
    \item We introduce \textbf{SemFM}, a semantic anchoring framework that integrates pretrained visual semantics with functional map–based surface correspondence.
    \item We propose a \textbf{semantic anchor selection method} that identifies mutually exclusive, high-confidence correspondences between semantically similar object regions.
    \item We develop an \textbf{efficient affordance transfer pipeline} that generalizes demonstrated interactions across object instances with low computational overhead.
\end{itemize}

\section{Related Work}\label{sec:related_work}

This work is most closely related to research on learning from demonstration and object-centric generalization for robotic manipulation, as well as representation learning for grasp selection and execution.

\textbf{Learning from Demonstration and Object-Centric Generalization.}
Learning from demonstration (LfD) and imitation learning \cite{ravichandar2020recent, Diff_templates} have emerged as central paradigms for enabling robots to acquire manipulation skills directly from human behavior, reducing the need for manually specified controllers or task models~\cite{lum2025crossinghumanrobotembodimentgap,chen2025tool, chen2026dexterousmanipulationpoliciesrgb, zhi2023learning}. A line of work focuses on object-centric formulations, where demonstrated motions or action primitives are associated with objects or object-relative frames, allowing interactions to be transferred to novel instances while respecting robot kinematic and feasibility constraints~\cite{dong2025jointflowtrajectoryoptimization,wen2022demonstrateoncecategorylevelmanipulation}. While effective for replaying demonstrated behaviors, such approaches often exhibit limited generalization across objects that are semantically similar but geometrically diverse. In parallel, large-scale efforts distill manipulation knowledge from diverse human video datasets to learn generalizable visuomotor policies that transfer across tasks, objects, and embodiments~\cite{kareer2024egomimic,bahl2022human,shi2025zeromimic}. Despite their success, these methods typically rely on end-to-end pipelines or implicit latent representations, making it difficult to explicitly control, interpret, or transfer localized interaction regions across geometries~\cite{herdliu2022,shaw2024demonstrating}.

\textbf{Grasp Selection and Representation Learning.}
Grasp selection and generalization have been extensively studied in robotics, with particular emphasis on achieving fast and reliable grasping in complex environments~\cite{wang2024graspness, rashidi2025gtg2}. Classical approaches formulate grasping as a geometric problem, where methods such as Grasp Pose Detection and the Grasp Pose Generator sample candidate grasp objects' geometry and evaluate them using analytic or learned quality metrics~\cite{gualtieri2016gpd,tenpas2017gpd}. More recent work incorporates learned representations to improve robustness and generalization, including point-based networks and large-scale grasp datasets that enable data-driven grasp evaluation across diverse object geometries~\cite{liang2019pointnetgpd, fang2020graspnet, sundermeyer2021contactgraspnet, zhang2024dexgraspnet2}. Beyond geometric scoring, other approaches explore learned grasp sampling directly, such as diffusion-based generative models that capture multimodal distributions of feasible grasps and scale to tens of millions of synthesized training examples~\cite{Zhang2024DexGraspDiffusionDU,graspgen2025}.

\section{Preliminaries: Functional Maps}\label{sec:preliminaries}

A core component of our method is the \emph{functional map} representation for computing dense correspondences between two shapes~\cite{ovsjanikov2012functional, Ovsjanikov2016ComputingAP}. Consider two object surfaces represented as triangle meshes $\mathcal{M}_1$ and $\mathcal{M}_2$. A classical correspondence is a pointwise map $T:\mathcal{M}_2 \rightarrow \mathcal{M}_1$ that matches each point (or vertex) on $\mathcal{M}_2$ to a point on $\mathcal{M}_1$. While intuitive, directly estimating $T$ is challenging in practice: pointwise maps are high-dimensional, sensitive to noise, and difficult to regularize.

Functional maps provide an alternative that is often more convenient for robotics and geometry processing. Instead of matching points directly, we match \emph{functions defined on the surface}. For example, a function can represent a region indicator, a heat distribution, or a semantic signal over the mesh. A pointwise correspondence $T$ induces a linear operator
\begin{align}
C : L^2(\mathcal{M}_2) \rightarrow L^2(\mathcal{M}_1), \qquad C(f) = f \circ T,
\end{align}
which transports functions from $\mathcal{M}_2$ to $\mathcal{M}_1$ by composition. Intuitively, if $f$ marks a region on $\mathcal{M}_2$ (e.g., a handle area), then $C(f)$ marks the corresponding region on $\mathcal{M}_1$. The operator $C$ is called a \emph{functional map}.

\begin{figure}[t]
    \centering
    \includegraphics[width=0.43\columnwidth]{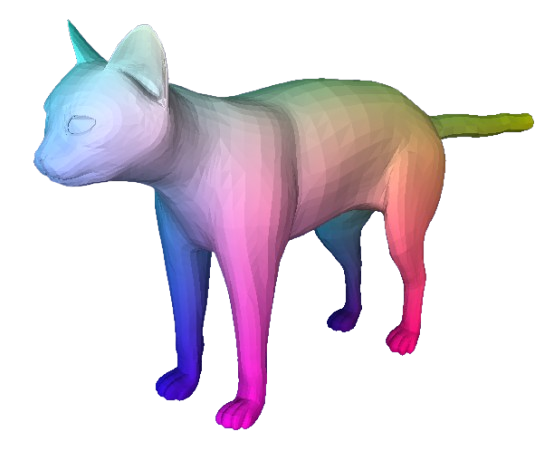}\hfill
    \includegraphics[width=0.41\columnwidth]{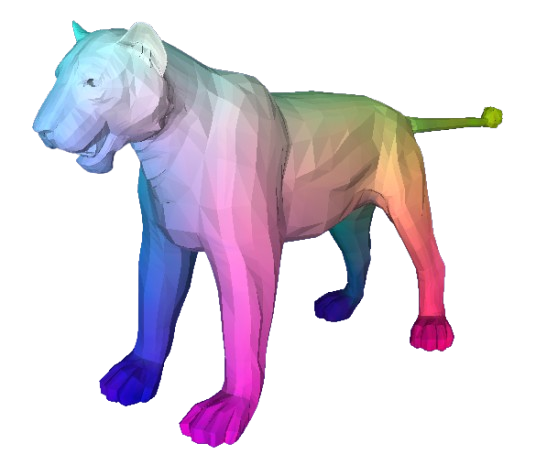}
    \caption{Illustration of functional maps. Given two related meshes (cat and tiger), functional maps recover a smooth correspondence that aligns semantically similar regions while maintaining spatial coherence over the surface.}
    \label{fig:fm_prelim}
    \vspace{-0.5em}
\end{figure}

To obtain a compact representation, functional maps are expressed in a low-dimensional basis, typically the first $k$ eigenfunctions of the Laplace--Beltrami operator~\cite{Reuter2006LaplaceBeltramiSA}. Let $\{\phi_i\}_{i=1}^k$ and $\{\psi_j\}_{j=1}^k$ denote the bases on $\mathcal{M}_1$ and $\mathcal{M}_2$, respectively. Any function can be approximated in this basis, and the functional map becomes a small matrix $\mathbf{C}\in\mathbb{R}^{k\times k}$ such that
\begin{align}
C(\psi_j) \approx \sum_{i=1}^k \mathbf{C}_{ij}\,\phi_i.
\end{align}
This spectral representation acts as a smoothness prior: low-frequency eigenfunctions capture large-scale structure, encouraging correspondences that vary smoothly over the surface. In practice, $\mathbf{C}$ is estimated by enforcing that certain functions (descriptors) correspond across shapes, resulting in a compact optimization problem. After computing $\mathbf{C}$, a pointwise map can be recovered by matching vertices in the induced functional embedding space, enabling dense region transfer between meshes. An illustrative example is shown in \cref{fig:fm_prelim}.

\section{Semantic Anchored Functional Map}\label{sec:METHOD}

\begin{figure*}[t]
    \centering
    \includegraphics[width=\textwidth]{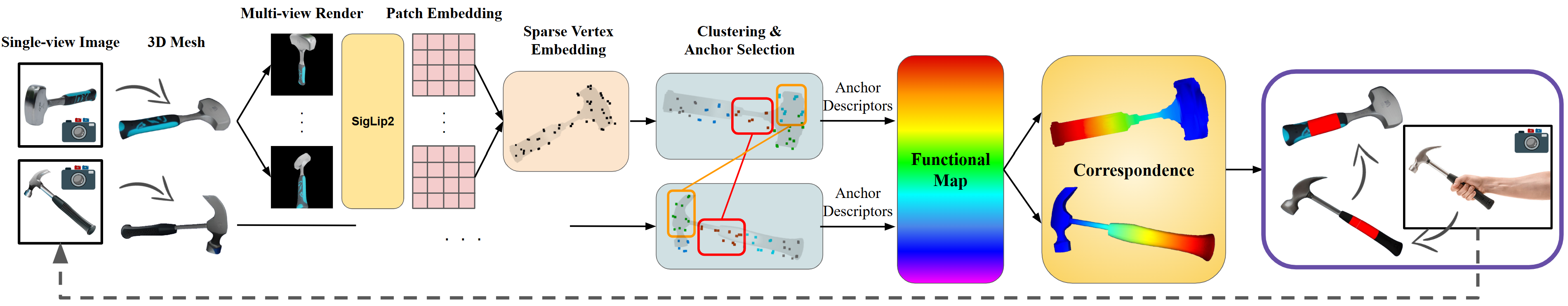}
    \caption{Overview of the proposed Semantic Anchored Functional Map pipeline. Given a single RGB observation of a demonstrated hand-object interaction and another object, we first construct coarse object meshes and extract the demonstrated affordance region. Semantic features are extracted using pretrained embeddings, lifted into 3D, and then used to identify corresponding anchor regions across objects, which constrain a functional map to produce smooth, dense correspondences. This then enables the affordance region to be transferred.}
    \label{fig:methodology_pipeline}
    \vspace{-2em}
\end{figure*}

\subsection{Problem Formulation}

Given an image of a demonstration $I_D$ that captures a human hand interacting with an object, and an image of observation $I_O$ that captures a different object, our goal is to determine where the human hand should interact with the observed object in 3D.

From these observations, we extract coarse triangular meshes representing the demonstrated object $M_1 = (V_1, F_1)$ and the observed object $M_2 = (V_2, F_2)$, where $V_1,V_2 \subset \mathbb{R}^3$ denotes the set of vertices and $F_1,F_2$ the set of faces. In addition, from the demonstrated interaction we obtain a 3D affordance region $A_D \subset V_1$ corresponding to the hand–object contact. Our objective is to identify a corresponding affordance region $A_O \subset V_2$ on the observed object that the hand should interact with, such that the transferred interaction respects \textbf{semantic continuity} between the two objects.

We formalize this requirement through a surface correspondence. A mapping $f_M : A_D \rightarrow A_O$ is said to respect semantic continuity if it is induced by a correspondence $f : V_1 \rightarrow V_2$ that is both geometrically smooth and semantically consistent across the two object surfaces.

Specifically, we assume the existence of a set of $\alpha$ semantically corresponding regions between $M_1$ and $M_2$, as
\begin{align}
\mathcal{K} = \{ (k_1^i, k_2^i) \mid i = 1, \dots, \alpha \},\label{eq:semantic-anchors}
\end{align}
where $k_1^i \subset V_1$ and $k_2^i \subset V_2$ are subsets of vertices representing semantically similar parts of the two objects. These regions serve as high-level \emph{semantic anchors} that constrain the correspondence. The goal is to compute a correspondence map $f : V_1 \rightarrow V_2$ such that
\begin{align}
f(k_1^i) \approx k_2^i, \quad \forall i = 1, \dots, \alpha,
\end{align}
where the approximation indicates consistency up to a small spatial or geodesic error. Beyond satisfying these anchor constraints, the correspondence $f$ is required to vary smoothly over the surface, preserving local neighborhood relationships between vertices.

Given such a semantically continuous correspondence, the affordance region on the observed object is defined as
\begin{align}
A_O = f(A_D), \label{eq:affordance_map}
\end{align}
which specifies the region on $M_2$ where the demonstrated interaction should be transferred. In contrast to traditional functional map approaches that rely primarily on intrinsic geometric structure, our method operates predominantly in the semantic domain. This enables meaningful correspondences between objects that may differ significantly in geometry but share semantically similar parts or components. An overview of the proposed framework is given in \cref{fig:methodology_pipeline}.

\subsection{Mesh and Affordance from Single-View Observation}
\label{sec:INFO_COL}

\begin{figure}[t]
    \centering
    \includegraphics[width=\columnwidth]{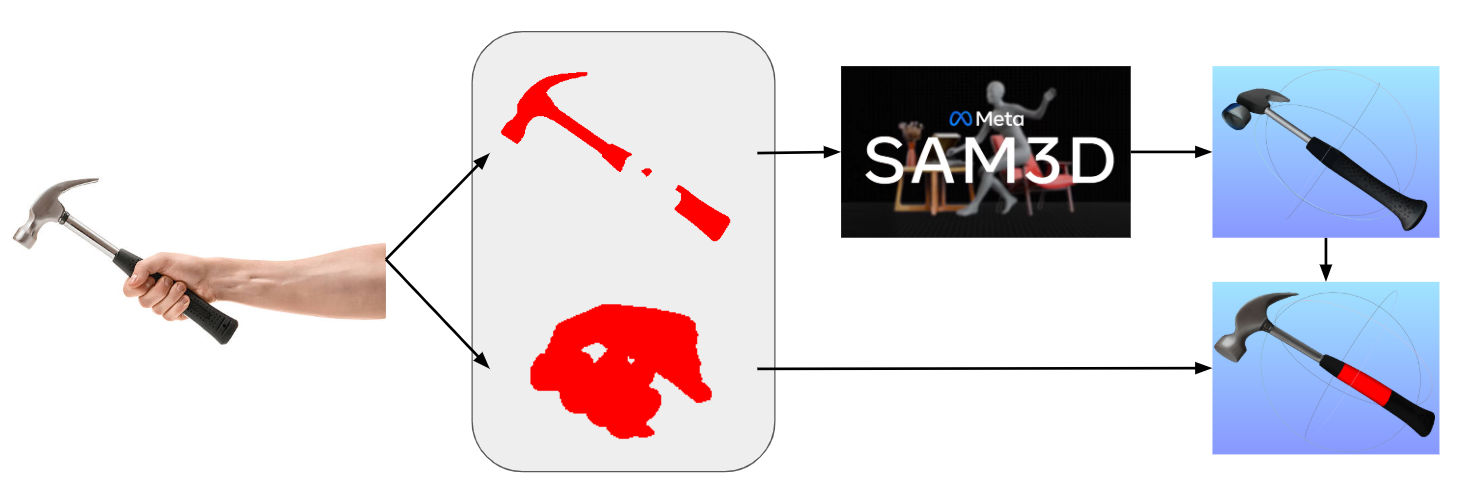}
    \caption{Pipeline acquiring coarse 3D mesh and affordance from single-view image. The mesh is inferred from an RGB image and object mask using SAM3D. The hand mask is lifted to 3D and intersected with the mesh to obtain the affordance region.}
    \label{fig:sam_3d_pipline}
\end{figure}

As shown in \cref{fig:sam_3d_pipline}, given an RGB image and an object mask, we reconstruct a triangular mesh using SAM3D~\cite{sam3dteam2025sam3d3dfyimages}. Notice our method does not require metrically accurate reconstruction and it suffices that the mesh preserves semantically meaningful part structure. We additionally extract a hand mask and lift it to 3D through the same pipeline, which serves as the input demonstrated affordance region $A_D$ in \cref{eq:affordance_map}. 

\subsection{3D Semantic Acquirement}\label{sec:3D_SEMANTIC}

Let $\mathcal{M} = (\mathcal{V}, \mathcal{F})$ denote a well-constructed triangular surface mesh representing a 3D object, where $\mathcal{V} \subset \mathbb{R}^3$ is the set of vertices and $\mathcal{F}$ the set of faces. We begin by defining a set of $N$ virtual cameras $\mathcal{C} = \{ \mathcal{C}_i \}_{i=1}^N$,
where each camera $\mathcal{C}_i$ is parameterized by intrinsic parameters $\mathbf{K}_i$ and extrinsic parameters $(\mathbf{R}_i, \mathbf{t}_i)$. Using a standard differentiable rendering operator $\mathcal{R}$, we generate a collection of RGB images
\begin{align}
\mathbf{I}_i = \mathcal{R}(\mathcal{M}; \mathcal{C}_i), \quad i = 1, \dots, N,
\end{align}
each corresponding to a distinct viewpoint of the mesh.

For each rendered image $\mathbf{I}_i$, we employ the pretrained \emph{SigLip2Vision} model ~\cite{Zhai2023SigmoidLF} to extract semantic features. Specifically, we extract the last hidden state of the model to obtain a per-patch embedding
\begin{align}
\mathbf{E}_i = \Phi(\mathbf{I}_i) = \{ \mathbf{e}_{i}^{(p)} \in \mathbb{R}^d \}_{p=1}^P.
\end{align}
where $P$ is the number of image patches and $d$ is the embedding dimension.

Each patch embedding $\mathbf{e}_{i}^{(p)}$ corresponds to a 2D image region, which can be associated with a 3D surface location via the known rendering process. Specifically, using the camera parameters and depth information from the rendering stage, we define a lifting operator
$\mathcal{L}_i : \mathbb{R}^2 \rightarrow \mathbb{R}^3$, which maps image coordinates to points on the mesh surface. Applying $\mathcal{L}_i$ to each patch yields a set of lifted semantic samples
\begin{align}
\mathcal{S}_i = \left\{ \left( \mathbf{x}_{i}^{(p)}, \mathbf{e}_{i}^{(p)} \right) \;\middle|\; \mathbf{x}_{i}^{(p)} \in \mathcal{M} \right\}.
\end{align}

Due to the patch-based nature of the image embeddings and potential view inconsistencies, the lifted semantic signals can be noisy and spatially redundant. To obtain a compact and robust 3D representation, we down-sample the mesh into a sparse point cloud of $M$ points,
$\mathcal{P} = \{ \mathbf{p}_j \in \mathbb{R}^3 \}_{j=1}^M$,
using uniform surface sampling. For each point $\mathbf{p}_j$, we aggregate nearby lifted embeddings by averaging the set of lifted samples within a spatial neighborhood of radius $\epsilon$.

As a result, we obtain a sparse semantic point cloud,
$\mathcal{P}_s = \left\{ (\mathbf{p}_j, \mathbf{s}_j) \right\}_{j=1}^M$,
where each point $\mathbf{p}_j$ is endowed with a semantic embedding $\mathbf{s}_j \in \mathbb{R}^d$ capturing the high-level meaning of its local geometric neighborhood. This representation serves as a compact and semantically enriched abstraction of the original 3D surface.

\subsection{Semantic Anchor Selection}\label{sec:SEM_ANCHOR}

\begin{figure}[t]
    \centering
    \includegraphics[width=\linewidth]{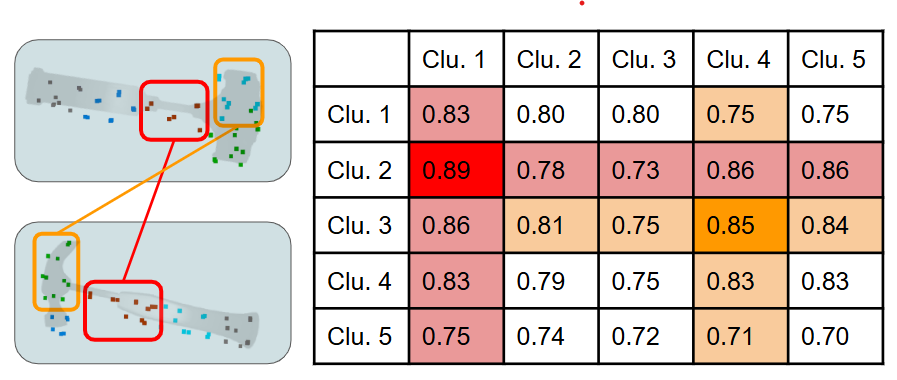}
    \caption{Example of semantic anchor selection. Each mesh is clustered into five regions. The top two \emph{mutually exclusive} cluster pairs are selected as semantic anchors based on cross-object similarity: the first anchor pairs cluster 2 of object 1 with cluster 1 of object 2 (similarity 0.89), and the second pairs cluster 3 of object 1 with cluster 4 of object 2 (similarity 0.85).}
    \label{fig:semantic-anchor}
\end{figure}

Given the sparse semantic point cloud 
$\mathcal{P}_s = \{ (\mathbf{p}_j, \mathbf{s}_j) \}_{j=1}^M$, our goal is to find a set of $\alpha$ semantic anchors as described in \cref{eq:semantic-anchors}. We first construct a weighted $k$-nearest neighbor (kNN) graph
$\mathcal{G} = (\mathcal{V}, \mathcal{E}, \mathbf{W})$,
where each vertex $v_j \in \mathcal{V}$ corresponds to a point $\mathbf{p}_j$, and edges $(v_j, v_\ell) \in \mathcal{E}$ connect each vertex to its $k$ nearest neighbors in Euclidean space. The edge weights are defined using the semantic embeddings as
\begin{align}
\mathbf{W}_{j\ell} = \exp\left( - \frac{\| \mathbf{s}_j - \mathbf{s}_\ell \|_2^2}{\sigma^2} \right),
\end{align}
yielding a graph that encodes semantic similarity over the surface.

We perform spectral clustering on $\mathcal{G}$ to partition the point cloud into $K$ disjoint semantic clusters, where each cluster $\mathcal{C}_c$ corresponds to a coherent semantic region of the object. For each cluster, we compute a cluster-level semantic embedding by averaging the embeddings of its constituent points:
\begin{align}
\bar{\mathbf{s}}_c = \frac{1}{|\mathcal{C}_c|} \sum_{(\mathbf{p}_j, \mathbf{s}_j) \in \mathcal{C}_c} \mathbf{s}_j.
\end{align}

Given two object meshes $\mathcal{M}^{(1)}$ and $\mathcal{M}^{(2)}$, we obtain corresponding sets of cluster embeddings $\{ \bar{\mathbf{s}}_c^{(1)} \}_{c=1}^{K_1}$ and $\{ \bar{\mathbf{s}}_{c'}^{(2)} \}_{c'=1}^{K_2}$.
To measure semantic compatibility between clusters across shapes, we compute a pairwise cosine similarity matrix
\begin{align}
\mathbf{S}_{cc'} = 
\frac{
\langle \bar{\mathbf{s}}_c^{(1)}, \bar{\mathbf{s}}_{c'}^{(2)} \rangle
}{
\| \bar{\mathbf{s}}_c^{(1)} \|_2 \, \| \bar{\mathbf{s}}_{c'}^{(2)} \|_2
},
\quad
\mathbf{S} \in \mathbb{R}^{K_1 \times K_2}.
\end{align}

From the similarity matrix $\mathbf{S}$, we select the top $\alpha$ cluster correspondences subject to a bijectivity constraint. Specifically, we seek a set of semantic anchor pairs
$\mathcal{A} = \{ (c, c') \}$,
that maximizes the total similarity score
\begin{align}
\max_{\mathcal{A}} \sum_{(c,c') \in \mathcal{A}} \mathbf{S}_{cc'},
\end{align}
such that $| \mathcal{A} | = \alpha$ and $c, c'$ exist only once $\forall (c, c') \in \mathcal{A}$. An example of anchor selection is shown in \cref{fig:semantic-anchor}. This constraint enforces a one-to-one correspondence between selected clusters, preventing multiple anchors of one object from mapping to the same cluster on the other object. The resulting anchor pairs serve as high-confidence semantic constraints for downstream shape matching.

\subsection{Functional Map with Semantic Anchors} \label{sec:FM_SA}

Given the set of $\alpha$ semantic anchor correspondences
$\mathcal{A} = \{ (c, c') \}$, we construct semantic descriptors defined over the full surface of each mesh. Let $\mathcal{M} = (\mathcal{V}, \mathcal{F})$ denote a mesh with vertex set $\mathcal{V} = \{ \mathbf{v}_i \}_{i=1}^{|\mathcal{V}|}$ and let
$\mathcal{P}_s = \{ (\mathbf{p}_j, \mathbf{s}_j) \}_{j=1}^M$ 
be its associated sparse semantic point cloud. For each vertex $\mathbf{v}_i$, we identify its closest point in the sparse point cloud,
\begin{align}
\pi(i) = \arg\min_{j} \| \mathbf{v}_i - \mathbf{p}_j \|_2.
\end{align}
A vertex $\mathbf{v}_i$ is said to belong to the anchoring region of semantic anchor $c$ if $\mathbf{p}_{\pi(i)} \in \mathcal{C}_c$, where $\mathcal{C}_c$ denotes the cluster associated with anchor $c$.

For each semantic anchor $c$, we define a binary indicator function over the mesh vertices,
\begin{align}
\mathbf{f}_c(i) =
\begin{cases}
1, & \text{if } \mathbf{v}_i \text{ belongs to anchor region } c, \\
0, & \text{otherwise}.
\end{cases}
\end{align}
These indicator functions provide an initial, sparse encoding of semantic regions but are typically noisy and discontinuous. To obtain smooth and spatially coherent descriptors, we apply heat diffusion over the mesh surface. Specifically, for each indicator function $\mathbf{f}_c$, we solve
\begin{align}
\tilde{\mathbf{f}}_c = \exp(-t \Delta) \mathbf{f}_c,
\end{align}
where $\Delta$ denotes the Laplace--Beltrami operator discretized on the mesh and $t > 0$ is a diffusion time parameter controlling the spatial extent of smoothing. The resulting functions are then normalized to unit $\ell_2$ norm.

We construct a semantic descriptor matrix for the mesh by stacking the diffused indicator functions:
\begin{align}
\mathbf{F} = 
\begin{bmatrix}
\tilde{\mathbf{f}}_1 & \tilde{\mathbf{f}}_2 & \cdots & \tilde{\mathbf{f}}_\alpha
\end{bmatrix}
\in \mathbb{R}^{|\mathcal{V}| \times \alpha}.
\end{align}
An analogous procedure is performed for both input meshes, yielding descriptor matrices $\mathbf{F}^{(1)}$ and $\mathbf{F}^{(2)}$. Using these semantic descriptors, we estimate a functional map between the two shapes. Let $\{ \phi_i^{(1)} \}_{i=1}^k$ and $\{ \phi_i^{(2)} \}_{i=1}^k$ denote the truncated Laplace--Beltrami eigenbases of the two meshes. The functional map $\mathbf{C} \in \mathbb{R}^{k \times k}$ is obtained by minimizing
\begin{align}
\min_{\mathbf{C}} 
\left\|
\mathbf{C} \langle \Phi^{(2)}, \mathbf{F}^{(2)} \rangle
-
\langle \Phi^{(1)}, \mathbf{F}^{(1)} \rangle
\right\|_F^2,
\end{align}
where $\Phi^{(i)}$ denotes the matrix of eigenfunctions evaluated at mesh vertices.

To improve accuracy and resolution, we further refine the functional map using the ZoomOut procedure~\cite{melzi2019zoomout}, progressively increasing the basis dimension while enforcing consistency between functional and pointwise correspondences. Finally, the refined functional map is converted into a vertex-level correspondence by mapping delta functions through the functional map and recovering pointwise matches via nearest-neighbor search in the spectral embedding space. This process yields a dense, semantically guided correspondence between the two meshes, combining high-level semantic anchors with intrinsic geometric structure.

\begin{figure*}[t]
    \centering
    \begin{subfigure}[b]{0.65\textwidth}
        \centering
        \includegraphics[width=\textwidth]{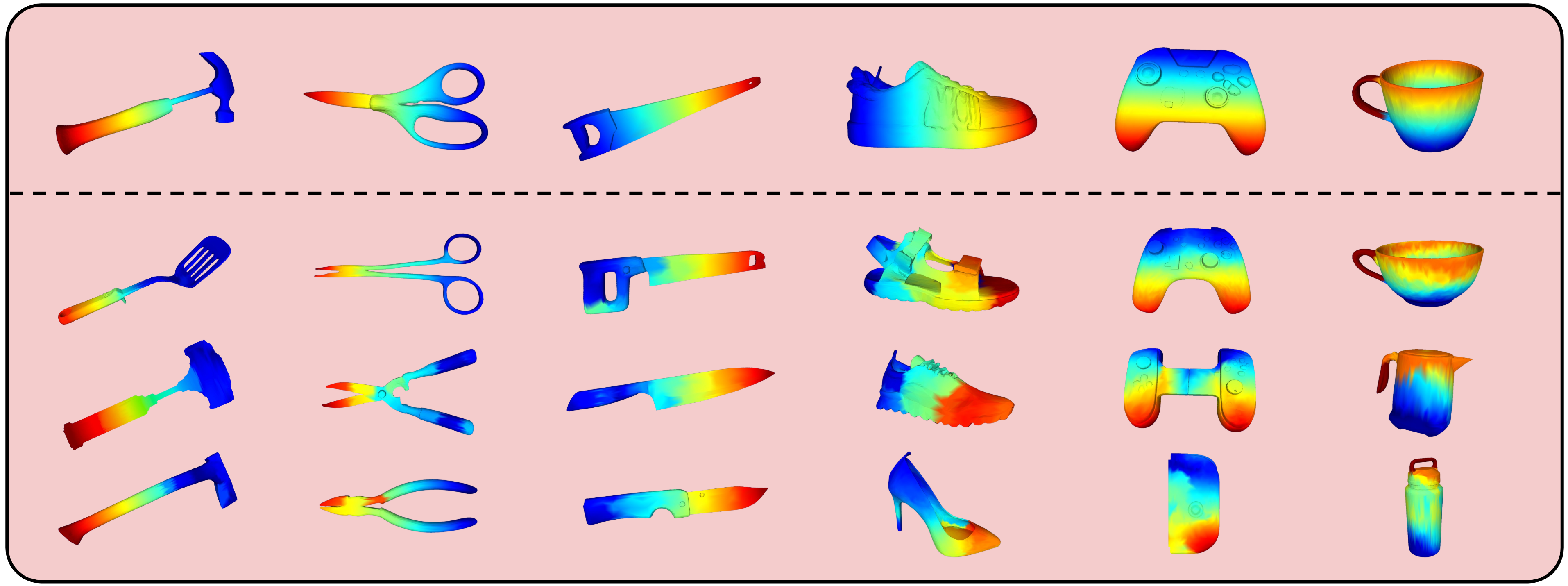}
        \caption{Illustration of transfer using SemFM for some selected objects in each category.
        Each column shows some transfer within each category. From left to right: Category 1, Category 6, Category 3, Category 2, Category 5, Category 2. The top row shows a reference coloring of the source object for each category, and the figures on the three rows of figures below are colorings of target object after the transfer from source.}
        \label{fig:experimental_results}
    \end{subfigure}
    \hfill
    \begin{subfigure}[b]{0.33\textwidth}
        \centering
        \includegraphics[width=\linewidth]{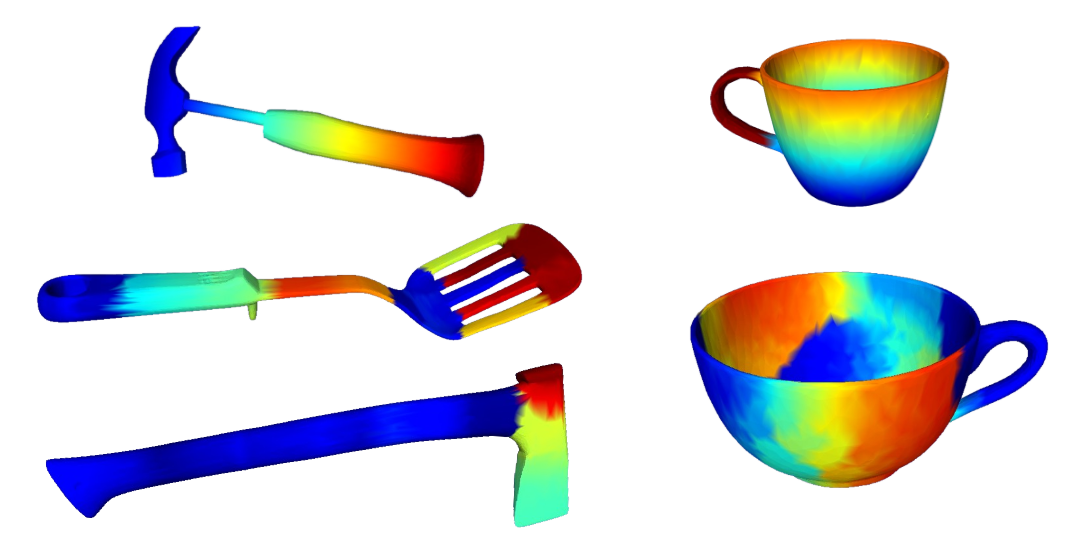}
        \caption{Affordance transfer examples using functional maps with WKS descriptors. The transferred regions often fail to align with semantically meaningful object parts and lack spatial consistency.}
        \label{fig:benchmark-defficiency-fm-wks}
    \end{subfigure}
    \caption{Qualitative results between our proposed SemFM against affordance transfer baselines.}
    \vspace{-1.75em}
\end{figure*}

\begin{figure}[t]
    \centering
    \includegraphics[width=\columnwidth]{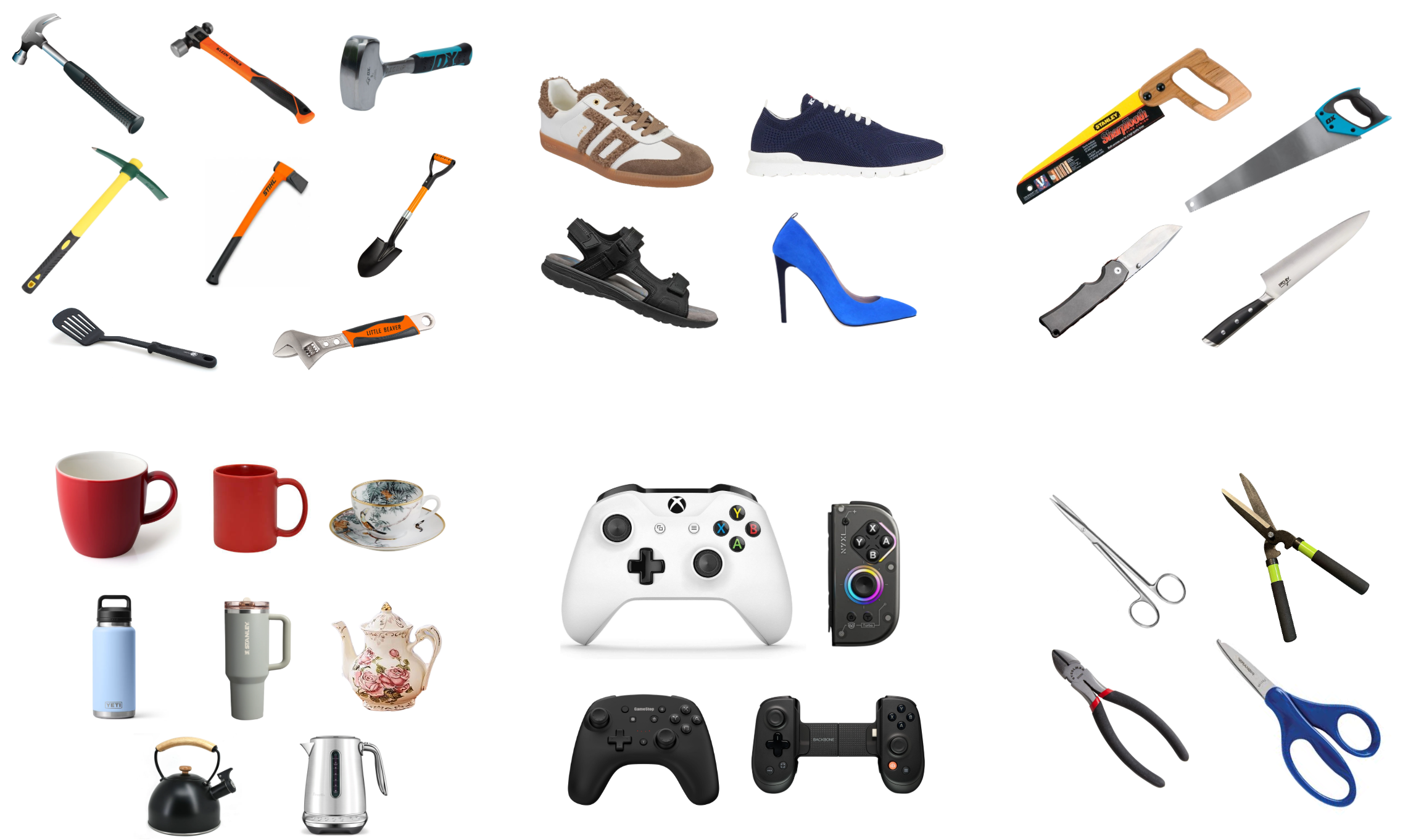}
    \caption{Visualization of selected six categories. From top-left to bottom-right: Category 1 containing tools with handles; Category 2 containing shoes; Category 3 containing cutting tools; Category 4 containing cups, kettles, and water bottles; Category 5 containing controllers; Category 6 containing scissor-type tools.}
    \label{fig:datasets}
\end{figure}

\section{Empirical Evaluations}\label{sec:EXPERIMENT}

We evaluate the proposed Semantic Anchored Functional Map framework on both synthetic and real-world examples. Our empirical study is designed to answer the following question:

\emph{Given a demonstrated affordance region on one object, can the proposed method transfer this region to a semantically corresponding and spatially coherent region on another object, even under significant geometric variation?}

To this end, we focus on affordance-aware region transfer across objects within the same semantic category, using both controlled synthetic data and real captured scenes.

\subsection{Evaluation on Synthetic Categories}

We first evaluate our method on synthetic categories designed to provide controlled semantic variation while maintaining consistent affordance definitions.

\paragraph{Category Construction}
We construct $6$ synthetic categories, each contains objects of different shapes while sharing some common semantic features, as shown in \cref{fig:datasets}. 
\begin{itemize}
    \item \textbf{Category 1:} tools with handles, such as hammers, pickaxes, and wrenches.
    \item \textbf{Category 2:} shoes such as sneakers, sandals, and high-heels. 
    \item \textbf{Category 3:} cutting tools such as saws and knives.  
    \item \textbf{Category 4:} water containers with handle, such as cups, kettles, and water bottles.  
    \item \textbf{Category 5:} controllers for gaming or televisions.  
    \item \textbf{Category 6:} scissor-type tools such as scissors, shears, and diagonal cutter.  
\end{itemize}

\paragraph{Synthetic Affordance Generation}
Using reference images and textual descriptions of the defined interaction regions, we employ a large language–vision model to generate images of human hands interacting with each object at the specified regions. These generated images provide explicit and consistent affordance demonstrations across objects. Applying the pipeline described in \cref{sec:INFO_COL}, we reconstruct object meshes and lift the hand–object interactions into 3D, yielding ground-truth affordance regions on each mesh.

\paragraph{Affordance Transfer Evaluation}

For each pair of meshes within the same category, we apply the proposed pipeline to compute a dense correspondence. Representative qualitative examples are shown in \cref{fig:experimental_results}. Using this correspondence, we transfer each affordance region from a source object to a target object. The transferred region is then compared against the ground-truth affordance region extracted from the synthetic interaction image. This procedure evaluates whether the transferred region aligns with the correct semantic part while maintaining spatial coherence on the target mesh. 

\subsection{Evaluation Metrics}

\paragraph{Accuracy.}
Affordance transfer accuracy is evaluated using the Intersection-over-Union (IoU) between the transferred affordance region and the ground-truth region on the target mesh. Let $A_{\text{pred}}$ and $A_{\text{gt}}$ denote the predicted and ground-truth affordance vertex sets, respectively. The IoU is defined as
\begin{align}
\text{IoU}(A_{\text{pred}}, A_{\text{gt}}) =
\frac{|A_{\text{pred}} \cap A_{\text{gt}}|}{|A_{\text{pred}} \cup A_{\text{gt}}|}.
\end{align}
For each category with $N$ objects, we evaluate all ordered pairs $(i, j)$, $i \neq j$, and report the average IoU over all affordance transfers:
\begin{align}
\overline{\text{IoU}} =
\frac{1}{N(N-1)} \sum_{i \neq j} 
\text{IoU}\!\left(A_{\text{pred}}^{(i \rightarrow j)}, A_{\text{gt}}^{(j)}\right).
\end{align}

\paragraph{Runtime.}
We additionally measure the end-to-end runtime required to transfer an affordance region from a source object to a target object. All timings exclude one-time preprocessing shared across methods (e.g., mesh reconstruction) and are averaged over all object pairs within each category using the same hardware setup. 

\subsection{Benchmarking against Other Methods}

\begin{figure}[t]
    \centering
    \begin{subfigure}[b]{0.60\linewidth}
        \centering
        \includegraphics[width=\columnwidth]{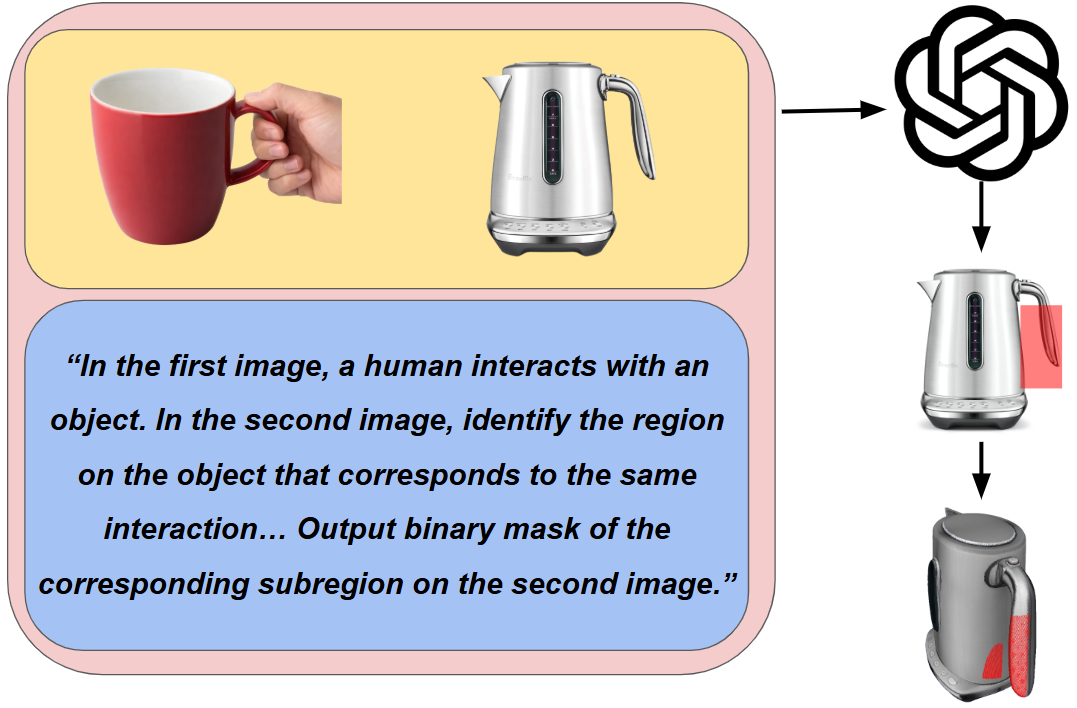}
        \caption{We compare against VLM-based baselines. Here, the generated mask is overlaid on top of the target image, highlighted in red, and the region in 3D is marked in red in the lower right.}
        \label{fig:gpt_benchmark_pipeline}
    \end{subfigure}
    \hfill
    \begin{subfigure}[b]{0.37\linewidth}
        \centering
        \includegraphics[width=\linewidth]{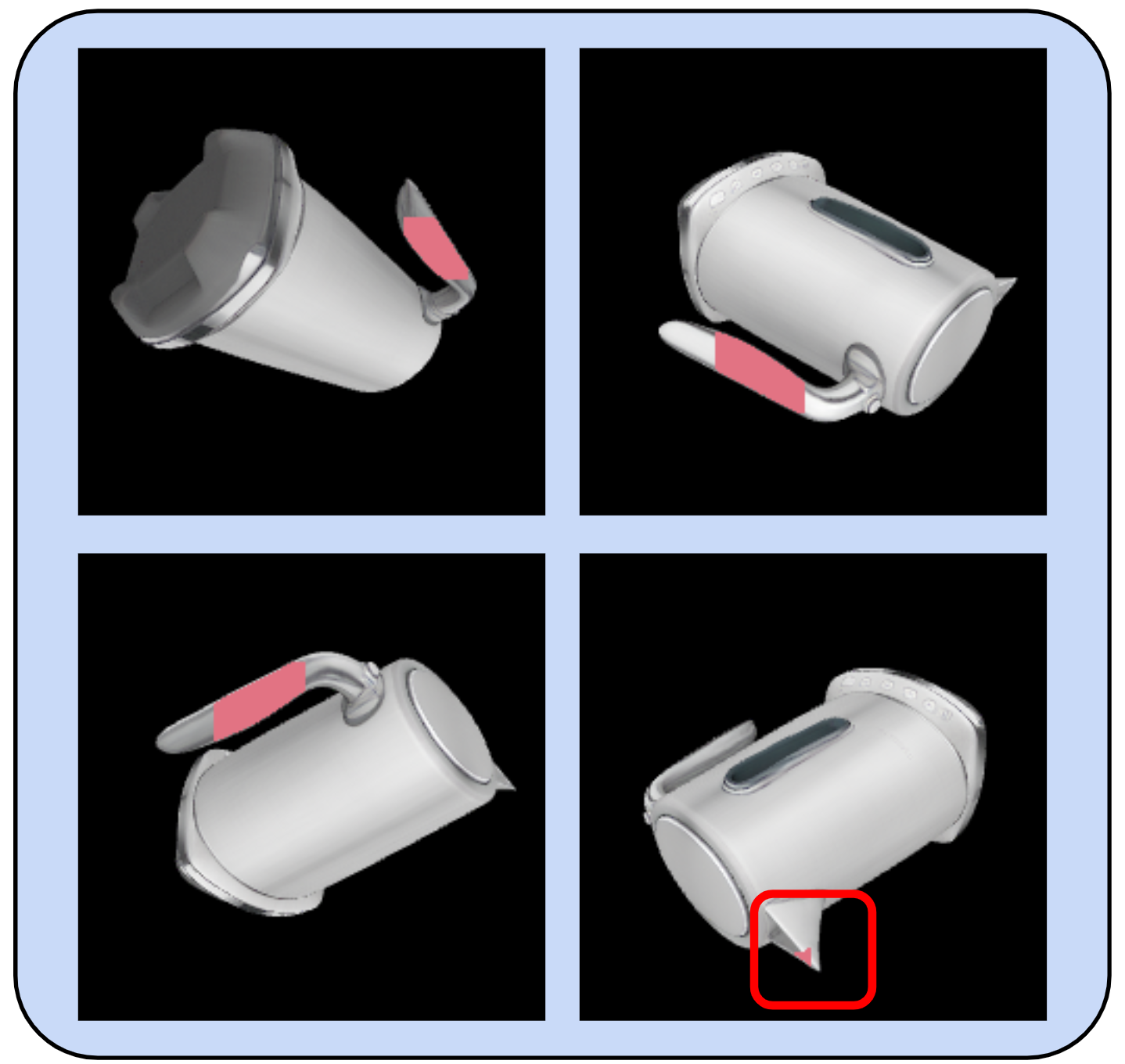}
        \caption{Example of affordance predictions (highlighted in red) by GPT-4o that are inconsistent across views.}
        \label{fig:benchmark-defficiency-vlm}
    \end{subfigure}
    \caption{GPT-4o comparison pipeline shown. Here, the results are slightly erroneous.}
    \label{fig:gpt-benchmark}
\end{figure}

We compare the proposed method against representative baselines for affordance transfer and shape correspondence, covering both geometry-driven and semantics-driven approaches.

\paragraph{Functional Maps with WKS Descriptors.}
As a geometry-based baseline, we evaluate a classical functional map pipeline using Wave Kernel Signature (WKS) descriptors~\cite{Aubry2011TheWK}. A functional map is first estimated from WKS features, after which vertex-level correspondences are recovered and used to transfer affordance regions. This method relies solely on intrinsic geometric structure and does not incorporate semantic information, serving as a representative baseline for traditional shape correspondence techniques.

\paragraph{Vision--Language Model (VLM) Baselines.}
We additionally benchmark against a vision-language–model-based affordance transfer pipeline, which has been adapted in many recent imitation learning work ~\cite{Chen2024VLMimicVL, Black20240AV, Huang2024ReKepSR, barron2025crossmodalinstructionsrobotmotion}. We choose GPT-4o as a representative state-of-the-art VLM due to its strong image understanding and semantic reasoning capabilities. Directly querying a VLM with 3D meshes is impractical due to input size constraints and limited 3D spatial reasoning. Instead, we adopt a multi-view, image-based strategy. Specifically, the source object is rendered from multiple viewpoints, with the affordance region annotated in image space. The VLM is prompted to identify the corresponding interaction region on rendered views of the target object. The predicted 2D regions are then lifted back to 3D using known camera parameters. We evaluate two settings: a single-view variant and a multi-view variant using six views. An example of the prompting and transfer process is shown in \cref{fig:gpt_benchmark_pipeline}. While increasing the number of views improves surface coverage, it also increases inference time and may introduce inconsistencies across views. These VLM-based baselines therefore serve as strong semantic references but incur substantial computational overhead.

\subsection{Performance Comparison on Synthetic Categories}

\begin{figure}[t]
    \centering
    \begin{subfigure}[b]{\columnwidth}
        \centering
        \includegraphics[width=\linewidth]{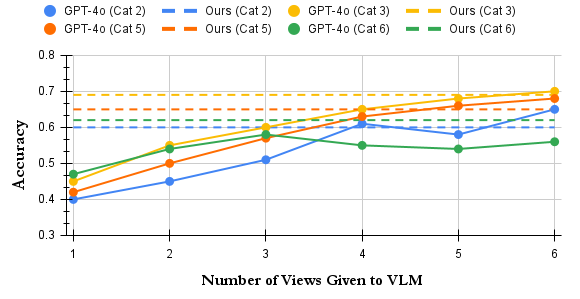}
        \caption{Affordance transfer IoU as a function of the number of rendered views provided to GPT-4o. With limited viewpoints, the VLM baseline underperforms our method due to incomplete surface coverage. Accuracy improves with additional views and becomes comparable in several categories at higher view counts.}
        \label{fig:transfer-ac-vs-views}
    \end{subfigure}
    \begin{subfigure}[b]{\columnwidth}
        \centering
        \includegraphics[width=\linewidth]{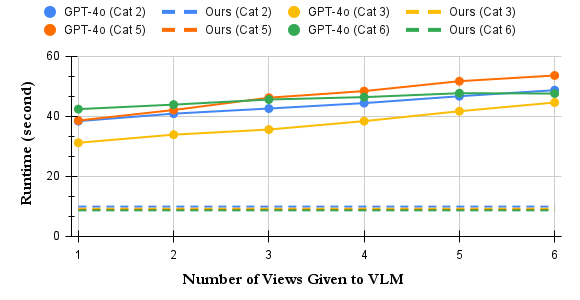}
        \caption{Average runtime per transfer as a function of the number of rendered views provided to GPT-4o. Increasing the number of views substantially increases inference time, whereas our method remains consistently efficient.}
        \label{fig:transfer-runtime-vs-views}
    \end{subfigure}

    \caption{Accuracy-runtime trade-off of multi-view VLM prompting. Solid curves show GPT-4o performance as the number of rendered views increases, while dotted horizontal lines indicate the performance of our method for each category.}
    \label{fig:transfer-performance-visual}
\end{figure}

\begin{figure*}[t]
    \centering
    \includegraphics[width=0.9\textwidth]{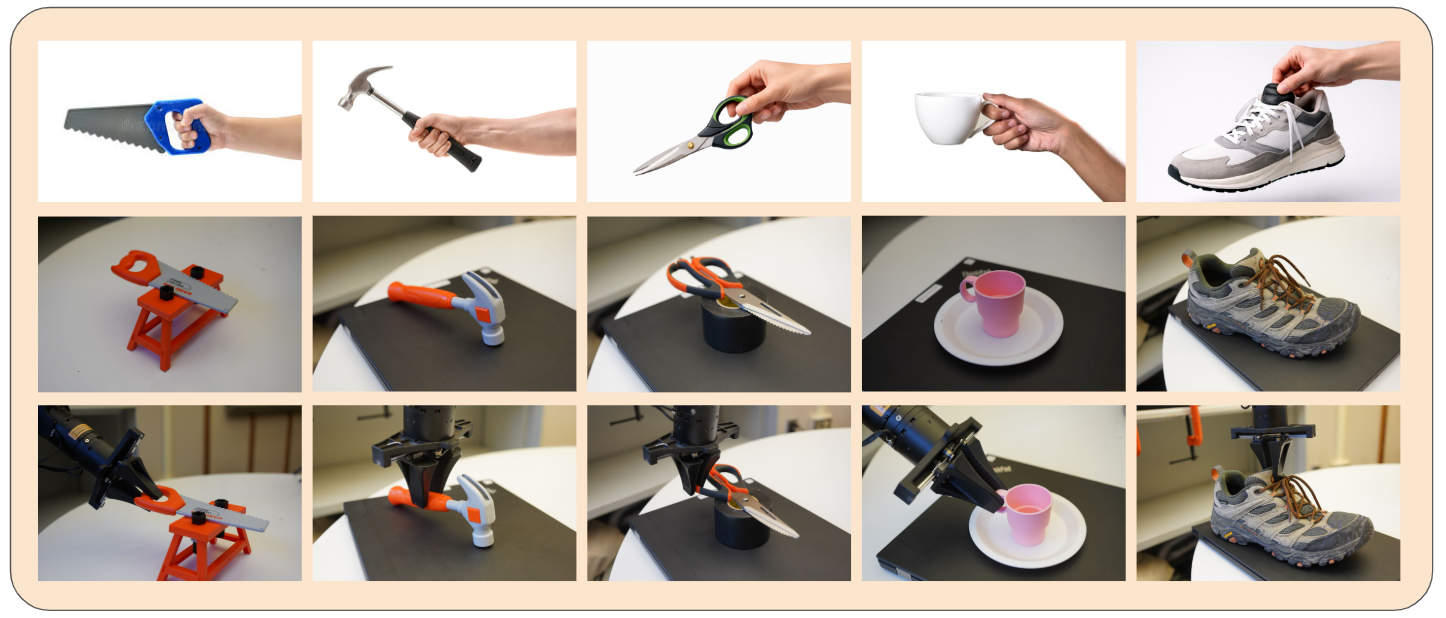}
    \caption{Illustration of Semantic-Anchored-Functional-Map performance in real-world. The leftmost image of the top row illustrates the setup used in real world; the rest of top row images show affordance demonstrations that are transferred from; the middle row images show the target real-world object to be transferred to; the last row images show the transferred affordance as a grasp executed by a robot arm with parallel-jaw gripper.}
    \label{fig:real-world-exp}
    \vspace{-1.5em}
\end{figure*}

We evaluate all methods on the synthetic categories described in the previous section. For each category, affordance transfer is performed across all ordered object pairs, and performance is measured using the averaged IoU metric. Some quantitative results are summarized in Tables~\ref{tab:quantitative_results} and visualized in \cref{fig:transfer-performance-visual}. Compared to our proposed method and the VLM-based baselines, the FM-WKS pipeline yields consistently low transfer accuracy. This failure is expected, as WKS descriptors are purely geometry-driven and do not capture semantic part similarity across geometrically diverse objects. As a result, FM-WKS often produces correspondences that are both semantically incorrect and spatially inconsistent, leading to poor affordance transfer quality (see \cref{fig:benchmark-defficiency-fm-wks}).

GPT-4o achieves strong affordance transfer accuracy in the multi-view setting, benefiting from its semantic reasoning capabilities. However, this improvement comes at a substantial computational cost: across all categories, GPT-4o requires significantly longer inference time than our method, and runtime increases monotonically with the number of views (see \cref{fig:transfer-performance-visual}). In addition, multi-view prompting may introduce view-wise inconsistencies and occasional hallucinated predictions, which can degrade mask quality after lifting to 3D (see \cref{fig:benchmark-defficiency-vlm}).

Overall, our method achieves affordance transfer accuracy comparable to multi-view VLM baselines while requiring substantially less computation time. Comparing to geometry-only methods, such as FM-WKS, semantic anchoring consistently produces transfers that are semantically meaningful and spatially coherent. These results highlight that combining semantic anchors with functional map propagation provides an effective balance between accuracy and efficiency, making the proposed approach well suited for real-world robotic pipelines that require fast perception-to-action loops.

\begin{table}[t]
\centering
\caption{Quantitative comparison on synthetic categories.}
\label{tab:quantitative_results}

\vspace{0.5em}

\begin{tabular}{lcccccc}
\toprule
Category & C.~1 & C.~2 & C.~3 & C.~4 & C.~5 & C.~6 \\
\midrule
FM-WKS        & 0.03 & 0.10 & 0.05 & 0.07 & 0.15 & 0.11 \\
VLM (1~POV)   & 0.48 & 0.40 & 0.45 & 0.45 & 0.42 & 0.47 \\
VLM (6~POV)   & 0.68 & \textbf{0.65} & \textbf{0.69} & \textbf{0.56} & \textbf{0.68} & 0.56 \\
Ours          & \textbf{0.70} & 0.60 & \textbf{0.69} & 0.53 & 0.65 & \textbf{0.62} \\
\bottomrule
\end{tabular}

\vspace{1em}

\begin{tabular}{lcccccc}
\toprule
Category & C.~1 & C.~2 & C.~3 & C.~4 & C.~5 & C.~6 \\
\midrule
VLM (1~POV)   & 31.7 & 38.4 & 31.2 & 35.7 & 38.6 & 42.4 \\
VLM (3~POV)   & 37.4 & 42.6 & 35.6 & 41.8 & 46.2 & 45.6 \\
VLM (6~POV)   & 45.2 & 48.7 & 44.6 & 46.1 & 53.6 & 47.6 \\
Ours          & \textbf{9.72} & \textbf{9.93} & \textbf{9.21} & \textbf{11.45} & \textbf{8.86} & \textbf{8.69} \\
\bottomrule
\end{tabular}

\vspace{0.5em}

{\small
\textbf{Top:} Affordance transfer accuracy (IoU; higher is better). \\
\textbf{Bottom:} Average runtime per transfer in seconds (lower is better).
}
\vspace{-1em}
\end{table}

\subsection{Real-World Robotic Experiments}

To demonstrate the practical applicability of the proposed method in robotic settings, we evaluate Semantic Anchored Functional Maps in real-world affordance transfer and grasp execution tasks. These experiments emphasize semantic generalization from arbitrary visual demonstrations to physically grounded robot actions. Unlike synthetic evaluations where dense ground truth affordance labels are available, quantitative evaluation of semantic correspondence in real-world settings is inherently ambiguous. Thus, we focus our real-world experiments on demonstrating system integration, qualitative semantic correctness, and practical feasibility, while reserving quantitative benchmarking for controlled synthetic categories. Note that affordance demonstrations for the source object are not restricted to the same sensing modality and may originate from web images or synthetic renderings. This decoupling allows affordance knowledge to be transferred across domains, highlighting the generalization capability of the proposed framework.

\paragraph{Mesh Reconstruction and Alignment}
Given a real-world RGB-D observation of the target object, we reconstruct a coarse mesh using SAM3D. Since single-view reconstruction does not recover metric scale, we align the reconstructed mesh to the observed depth data using rigid alignment, resulting in metrically consistency for downstream planning. 

\subsubsection{Affordance Transfer in the Real World}
Using the reconstructed and aligned target mesh, we apply the proposed semantic anchoring and functional map pipeline to transfer the affordance region from the source object to the target object. The resulting affordance region is defined as a subset of vertices on the target mesh, representing semantically appropriate interaction areas (e.g., handle regions).

\subsubsection{Grasp Generation and Execution}
To convert the transferred affordance region into a physically feasible grasp, we employ a Grasp Pose Generator (GPG) operating on the scaled target mesh. Candidate grasps are sampled within the predicted affordance region and ranked based on standard geometric and collision-based criteria. The highest-scoring grasp is selected, converted into a robot end-effector pose, and executed using inverse kinematics and a motion planner.

Selected examples of transfer from source objects to target objects and the resulting grasp executions are shown in \cref{fig:real-world-exp}. These experiments illustrate that the proposed method can transfer demonstrated affordance regions from visual observations to real-world robotic manipulation. By combining semantic correspondence with classical grasp planning, the framework supports consistent generalization across object instances while remaining compatible with standard robotic perception and control pipelines.

\section{Conclusion and Future Work}

In this paper, we presented \emph{Semantic Anchored Functional Maps}, an efficient framework for transferring affordances across object instances by combining high-level semantic alignment with smooth geometric correspondence. By anchoring correspondence at semantically meaningful regions and propagating these constraints using functional maps, our approach enables accurate affordance transfer between geometrically diverse objects while operating at significantly lower computational cost than VLM-based pipelines. Extensive evaluation on synthetic benchmarks and real-world experiments demonstrates that our method achieves a favorable balance between semantic accuracy, efficiency, and practical applicability, making it well suited for learning-from-demonstration scenarios in robotics. Avenues for future research include: (1) integrating support for depth information if provided; (2) integrating temporal information, where the affordances are continuously refined with additional frames of the target object.

\bibliographystyle{ieeetr} % We choose the "plain" reference style
\bibliography{bib}
\end{document}